\documentclass[10pt,twocolumn,letterpaper]{article}

\usepackage{iccv}

\usepackage{times}
\usepackage{graphicx}
\usepackage{amsmath}
\usepackage{amssymb}
\usepackage{booktabs}

\usepackage{url}            % simple URL typesetting
\usepackage{booktabs}       % professional-quality tables
\usepackage{amsfonts}       % blackboard math symbols
\usepackage{nicefrac}       % compact symbols for 1/2, etc.
\usepackage{microtype}      % microtypography
\usepackage{epsfig}
\usepackage{colortbl}
\usepackage{multicol}
\usepackage{multirow}
\usepackage{float}
\usepackage{multirow}
\usepackage{url}
\usepackage{makecell}
\usepackage{subcaption}
\usepackage{cuted}
\usepackage[font={small}]{caption}
\usepackage{pifont}
\usepackage{lipsum} 
\usepackage[dvipsnames,table,xcdraw]{xcolor}
\usepackage{caption}
\usepackage[misc]{ifsym}

\usepackage[accsupp]{axessibility}

\usepackage[pagebackref=true,breaklinks=true,letterpaper=true,colorlinks,bookmarks=false]{hyperref}

\iccvfinalcopy % *** Uncomment this line for the final submission

\def\iccvPaperID{5957} % *** Enter the ICCV Paper ID here

\ificcvfinal\fi

\newcommand\blfootnote[1]{%
  \begingroup
  \renewcommand\thefootnote{}\footnote{#1}%
  \addtocounter{footnote}{-1}%
  \endgroup
}

\begin{document}

\title{3DHumanGAN: 3D-Aware Human Image Generation with 3D Pose Mapping}  % **** Enter the paper title here

\author{Zhuoqian Yang$^{1, 2 \dagger}$ \qquad Shikai Li$^{1}$ \qquad Wayne Wu\textsuperscript{1 \Letter} \qquad Bo Dai$^{1}$  \\
$^{1}$ Shanghai AI Laboratory \hspace{10pt}
$^{2}$ School of Computer and Communication Sciences, EPFL \\
% \vspace{-5mm}
{\tt\small zhuoqian.yang@epfl.ch}\qquad
{\tt\small lishikai@pjlab.org.cn}\qquad
\\
{\tt\small wuwenyan0503@gmail.com}\qquad
{\tt\small daibo@pjlab.org.cn}
% \vspace{-10mm}
}

\twocolumn[{%
\renewcommand\twocolumn[1][]{#1}%
\maketitle
\begin{center}
    \vspace{-7mm}
    \includegraphics[width=1\textwidth]{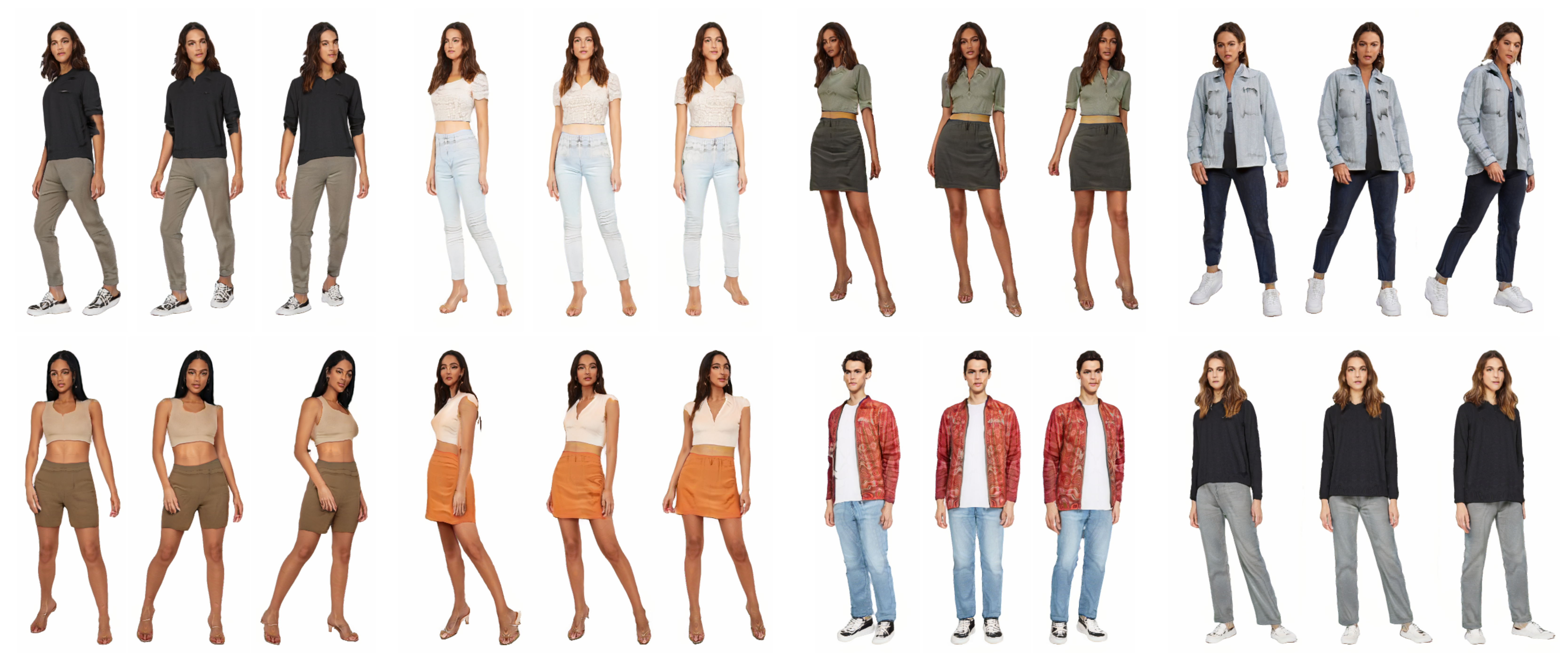}
    \vspace{-5mm}
    \captionof{figure}{\textbf{3DHumanGAN.} View-consistent full-body human images generated by our 3D-aware generative adversarial network (GAN). We show random sampled eight humans, each of them is in three view-angles.
    % \textbf{View-consistent full-body human image generation.} We present a novel 3D-aware generator 
    }
    \vspace{-1mm}
    \label{fig:teaser}
\end{center}
}]

\blfootnote{$\dagger$ Work done as research engineer at Shanghai AI Laboratory.}
\blfootnote{Project page: \url{https://3dhumangan.github.io/}}

\begin{abstract}
\vspace{-3mm}
We present \textbf{3DHumanGAN}, a 3D-aware generative adversarial network that synthesizes photo-like images of full-body humans with consistent appearances under different view-angles and body-poses. To tackle the representational and computational challenges in synthesizing the articulated structure of human bodies, we propose a novel generator architecture in which a 2D convolutional backbone is modulated by a 3D pose mapping network. The 3D pose mapping network is formulated as a renderable implicit function conditioned on a posed 3D human mesh. This design has several merits: i) it leverages the strength of 2D GANs to produce high-quality images; ii) it generates consistent images under varying view-angles and poses; iii) the model can incorporate the 3D human prior and enable pose conditioning.
% The 3D pose mapping network is formulated as a renderable implicit function which is inherently view-consistent.
%
% \wayne{may also add the description of the appearance mapping network}
%
% To preserve this consistency during the 2D synthesis process, we propose a network design with two key aspects: 1) our backbone is a pixel-wise independent convolutional network which avoids positional reference and promotes equalvariance; 2) modulation from the pose mapping network is passed into the backbone by means of spatial adaptive batch normalization instead of instance normalization so that underlying structure is not impaired. 
Our model is adversarially learned from a collection of web images needless of manual annotation.
% Project page: \url{https://3dhumangan.github.io/}.

% \footnote{Project page: \url{https://3dhumangan.github.io/}
% Code: \url{https://github.com/3dhumangan/3DHumanGAN}}.

\end{abstract}

\thispagestyle{empty}

\section{Introduction}

Human image generation is a long-standing topic in computer vision and graphics with applications across multiple areas of interest including movie production, social networking and e-commerce. Compared to physically-based methods, data-driven approaches are preferred due to the photolikeness of their results, versatility and ease of use \cite{tewari2020state}. In this work, we are interested in synthesizing full-body human images with a 3D-aware generative adversarial network (GAN) that produces appearance-consistent images under different view-angles and body-poses.

Rapid developments have been seen in using 3D-aware GANs to generate view-consistent images of human faces \cite{niemeyer2021giraffe, schwarz2020graf, chan2021pi, chan2021efficient, gu2021stylenerf, zhou2021CIPS3D, hong2021headnerf}. However, these methods have limited capacity when dealt with complex and articulated objects such as human bodies. To begin with, methods based solely on neural volume rendering \cite{chan2021pi, niemeyer2021giraffe, schwarz2020graf} are too memory inefficient. Rendering human bodies requires a volumetric representation that is much more dense than that of faces, which makes it computationally infeasible. A line of work improves the computational efficiency and rendering quality of 3D-aware GANs by refining the rendered output with a convolutional neural network \cite{zhou2021CIPS3D, gu2021stylenerf, chan2021efficient, hong2021headnerf}. 
% We argue that this feed-forward approach is not well-suited for generating full-body human images, since the 3D representation is tasked with simultaneously modeling the articulated geometry and the appearance of human bodies, which poses great challenges on representational capacity, while the capacity of the 2D-network is not fully utilized. 
However, we argue that this method is not optimal for generating full-body human images. This is because the 3D representation has to capture both the shape and the appearance of human bodies at the same time, which requires a high level of representational capacity. Meanwhile, the potential of the 2D-network is not fully exploited.

\begin{figure*}[!t]
\centering
\begin{minipage}[b]{0.72\textwidth}
\includegraphics[width=\linewidth]{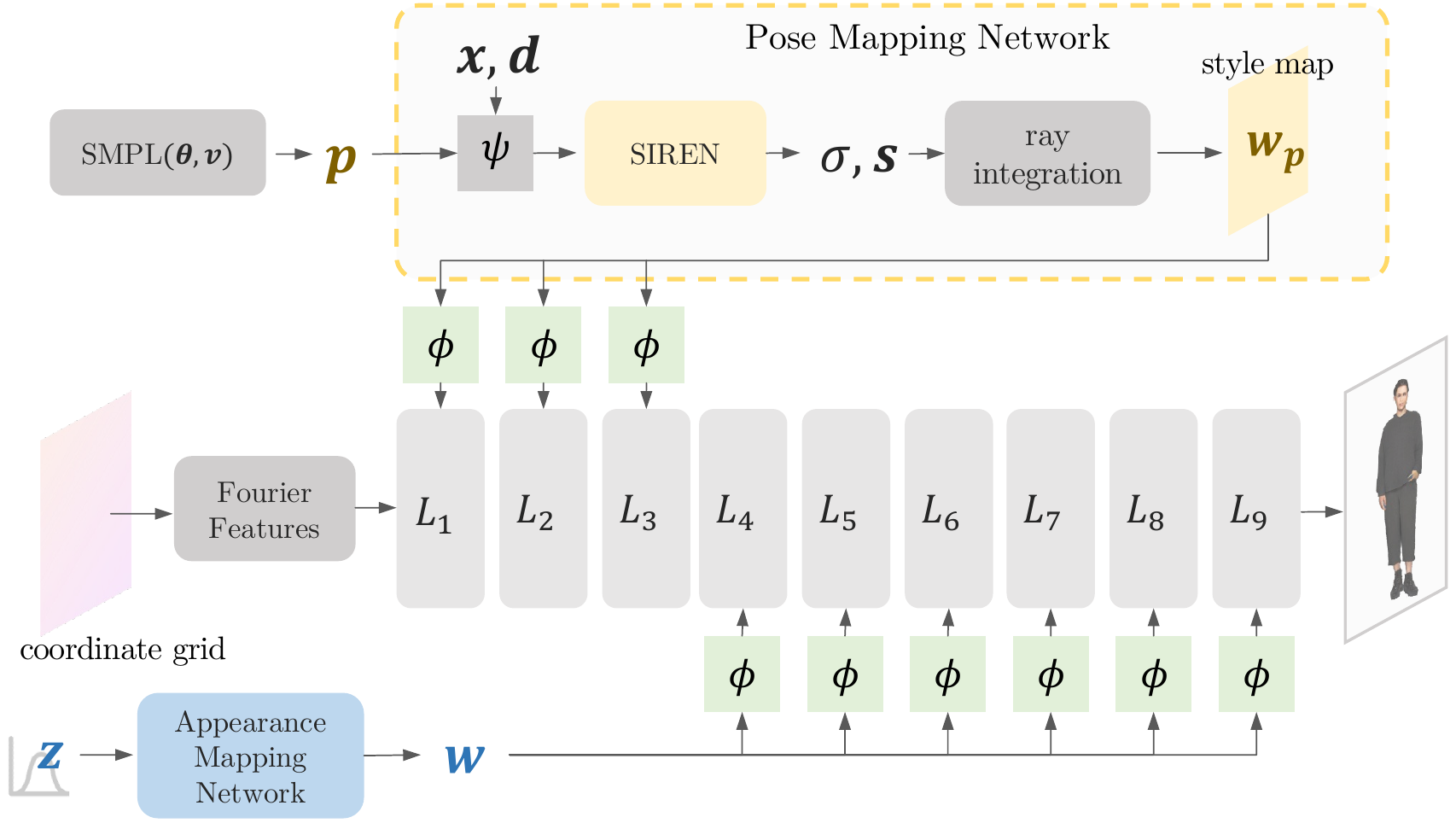}
\vspace{-2ex}
\subcaption{\small{Generator Architecture}}
\label{fig:gen}
\end{minipage}%
\begin{minipage}[b]{0.28\textwidth}
\includegraphics[width=\linewidth]{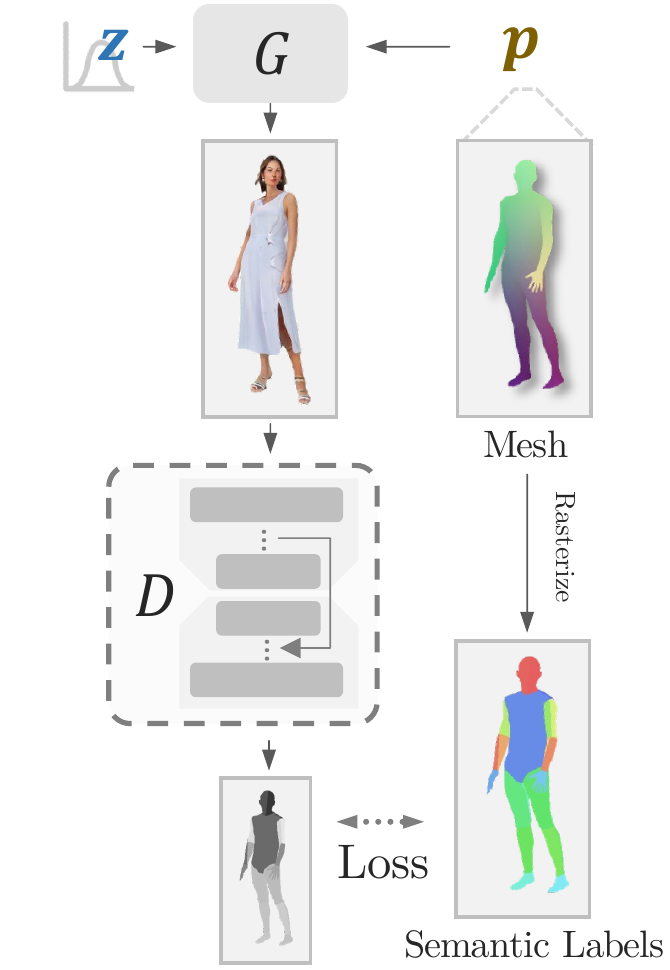}
\vspace{-2ex}
\subcaption{\small{Training Pipeline}}
\label{fig:train}
\end{minipage}%
\vspace{-2ex}
\caption{\textbf{Architecture.} \textbf{(a)} Our generator uses a pixel-wise independent convolutional backbone  modulated by a 3D pose mapping network  and an appearance mapping network. \textbf{(b)} Our model is supervised with a segmentation-based GAN loss where the semantic labels are rasterized from the conditioning mesh. }
\end{figure*}

As depicted in Figure \ref{fig:gen}, our work introduces a novel generator architecture in which a 2D convolutional backbone is modulated by a 3D pose mapping network. This design is motivated by the observation that in a StyleGAN2~\cite{stylegan2} model trained on human images certain layers of styles correlate strongly with the pose of the generated human \cite{fu2022styleganhuman} while others correlate more apparently with the appearance.
% \shikai{This design is motivated by \cite{fu2022styleganhuman}. As shown in Fig. \ref{fig:stylemix}, the low- and high-level styles are highly correlated with the pose and appearance of generated human respectively.}
The 3D Pose Mapping Network is formulated as a renderable implicit function conditioned on a posed 3D human mesh derived with a parametric model \cite{SMPL:2015}. In this way, the 3D representation handles the simplified task of parsing a geometric prior. As an additional benefit of explicitly conditioning on posed human mesh, the pose of the generated human can be specified. The output of the 3D pose mapping network is used to render a 2D low-resolution style map through ray integration \cite{mildenhall2020nerf}. The style map is passed into the first few layers of our backbone network. The appearance of the generated human is controlled by the Appearance Mapping Network. It is formulated as an MLP following a common practice in style-based generators \cite{stylegan, stylegan2}. For the network to learn to parse the 3D geometric prior, we use a segmentation-based GAN loss \cite{oasis2021} calculated using a U-Net \cite{ronneberger2015unet} discriminator. This design enables the network to establish one-to-many mapping from 3D geometry to synthesized 2D textures using only collections of single-view 2D photographs without manual annotations.

% Since the 3D pose mapping network and the appearance mapping network function independently, 
% highlight {pose control}
Traditional CNN generator networks with $3\times3$ convolution and progressive upsampling are not equivariant and demonstrate inconsistency under geometric transformations \cite{karras2021alias}. In our case, the appearance of the generated human may change when pose and view-angle vary. To preserve consistency, we propose a network design with two key aspects: 1) Our backbone network is built entirely with $1 \times 1$ convolutions. This helps eliminate positional reference and promotes equalvariance. 2) Modulation from the pose mapping network is passed into the backbone by means of spatial adaptive batch normalization instead of the commonly used instance
normalization \cite{stylegan2, park2019spade}, so that underlying structure from the geometric information parsed by our 3D pose mapping network is preserved.
% \bdai{We will show in the experiments that these design choices are crucial for synthesizing human images of consistent appearance under varying pose and view angles.(this sentence didn't say anything informative. you can remove it)}

Our contributions can be summarized as follows:
1) We propose a 2D-3D hybrid generator which is both efficient and expressive. The model is supervised with segmentation-based GAN loss which helps establish a mapping between 3D coordinates and 2D human body semantics.
% The 3D mapping network is inherently view-consistent and the 2D backbone is carefully designed to preserve the consistency of appearance when pose and view-angle varies.
2) Our generator is carefully designed to preserve the consistency of appearance when pose and view-angle vary.
3) Our work achieves state-of-the-art fidelity of 3D-aware generation of full-body human images.

\section{Related Works}

\noindent\textbf{3D-Aware Image Generation}.  Generative Adversarial Networks \cite{goodfellow2014generative} are able to generate images of human/animal faces, cars, indoor and natural scenes in eye-deceiving photorealisticness \cite{stylegan,stylegan2,karras2021alias}. This motivates researchers to explore adversarial learning's potential in achieving 3D-aware image synthesis. Earlier efforts devised different network structures to process meshes \cite{liao2020towards, szabo2019unsupervised}, voxels \cite{zhu2018visual, gadelha20173d, henzler2019escaping} and block-based representations \cite{hao2021gancraft, liu2020neural}. These methods suffer from the common problem of insufficient 3D inductive bias \cite{chan2021efficient}. Neural implicit representations rendered via ray integration have been found to be an effective representation for data-driven 3D synthesis \cite{lombardi2019neural}. Progress was gained in generating objects with simple structure such as human faces \cite{chan2021pi, niemeyer2021giraffe, schwarz2020graf}. These approaches are burdened by heavy computational cost and limited to generating low-resolution images. Several works propose to remedy this by upsampling the rendered results with a 2D convolutional network and achieved impressive quality \cite{zhou2021CIPS3D, gu2021stylenerf, chan2021efficient, hong2021headnerf}, yet still not ready for the challenging 3d-aware human image generation. This is because the complexity of the articulated structure of human bodies challenges the 3D representations used by these works. 
% \shikai{Should we add more details here for the motivation?}

\noindent\textbf{Human Image Generation}. Previous works on person image generation focus on re-synthesizing the reference image, such as rendering humans in a novel pose~\cite{ma2017pose,PATN,posewithstyle,cheng2022generalizable, sanyal2021learning}, with different garments~\cite{han2018viton,tryongan} or with text~\cite{jiang2022text2human}. Earlier methods condition sparse key-points~\cite{ma2017pose,ma2018disentangled,PATN,GFLA,styleposegan} or semantic maps~\cite{dong2018soft,han2019clothflow,transmomo2020, grigorev2021stylepeople} to manipulate the 2D images. To better preserve the appearance consistency of the source, surface-based methods~\cite{li2019dense,lwb2019,posewithstyle} are introduced to establish the dense correspondences between pixel and human prior surface, \ie , DensePose~\cite{Guler2018DensePose} and SMPL~\cite{SMPL:2015}. A limited amount of work focuses on human synthesis without a reference image, where \cite{grigorev2021stylepeople} and \cite{sarkar2021humangan} map the appearance into a gaussian distribution to enable free sampling of identities. Inspired by the success of StyleGAN~\cite{stylegan,stylegan2,karras2021alias} in unconditional generation, several works~\cite{fu2022styleganhuman, fruhstuck2022insetgan} explore the capacity of the model for full-body synthesis from different aspects. Specifically, StyleGAN-Human~\cite{fu2022styleganhuman} takes a data-centric perspective toward human generation and discusses the effect of manipulated style embedding on human images. InsetGAN~\cite{fruhstuck2022insetgan} proposes a multi-GAN optimization framework to synthesize more plausible-looking humans. All methods mentioned above are 2D methods.

Three concurrent works~\cite{bergman2022generative,zhang2022avatargen,hong2022eva3d} tackle 3D-aware human image generation based on different 3D representations. \cite{bergman2022generative} and \cite{zhang2022avatargen} render a low-resolution image with a 3D triplane representation \cite{chan2021efficient} and uses a 2D convolutional network to enlarge the image in a feed-forward manner. \cite{hong2022eva3d} render images directly with an efficient compositional 3D neural field.
Our work is different from these works in the following aspects: 1) We use an equivariant 2D generator modulated by 3D human body prior instead of directly rendering the image from a 3D representation. 2) We use segmentation-based adversarial supervision which encourages not only fidelity but also semantic correspondence.
% \wayne{need to revise}

% \zhuoqian{Put this? As a result, we are able to generate images with a resolution higher than that of \cite{bergman2022generative} and \cite{zhang2022avatargen} and more diverse appearances than that of \cite{hong2022eva3d}}
% \shikai{should we add two papers of 3d-aware human gan?}

\section{Generator Architecture}

\noindent We build a generative adversarial network that synthesizes 3D-aware full-body human images with specified pose $\mathbf{p}$ and view-angle $\mathbf{v}$. Pose is specified as a set of coordinates of the vertices in a posed body mesh $\mathbf{p} = \{\mathbf{v}_i \in \mathbb{R}^{3}\}_{i=1..6890}$ derived using a parametric human body model SMPL \cite{SMPL:2015} $\mathbf{p} = \text{SMPL}(\boldsymbol{\theta}, \mathbf{v})$. $\boldsymbol{\theta} \in \mathbb{R}^{K \times 4}$ denotes the rotation at each body joint in the quaternion format. The appearance of the generated human is randomly sampled, represented as a vector $\mathbf{z} \in \mathbb{R}^{N_z}$, but required to stay consistent when pose and view-angle vary. To this end, we design a 2D-3D-hybrid generator architecture and a training strategy to achieve partially conditioning on pose. 

\begin{figure}
    \centering
    \includegraphics[width=0.8\linewidth]{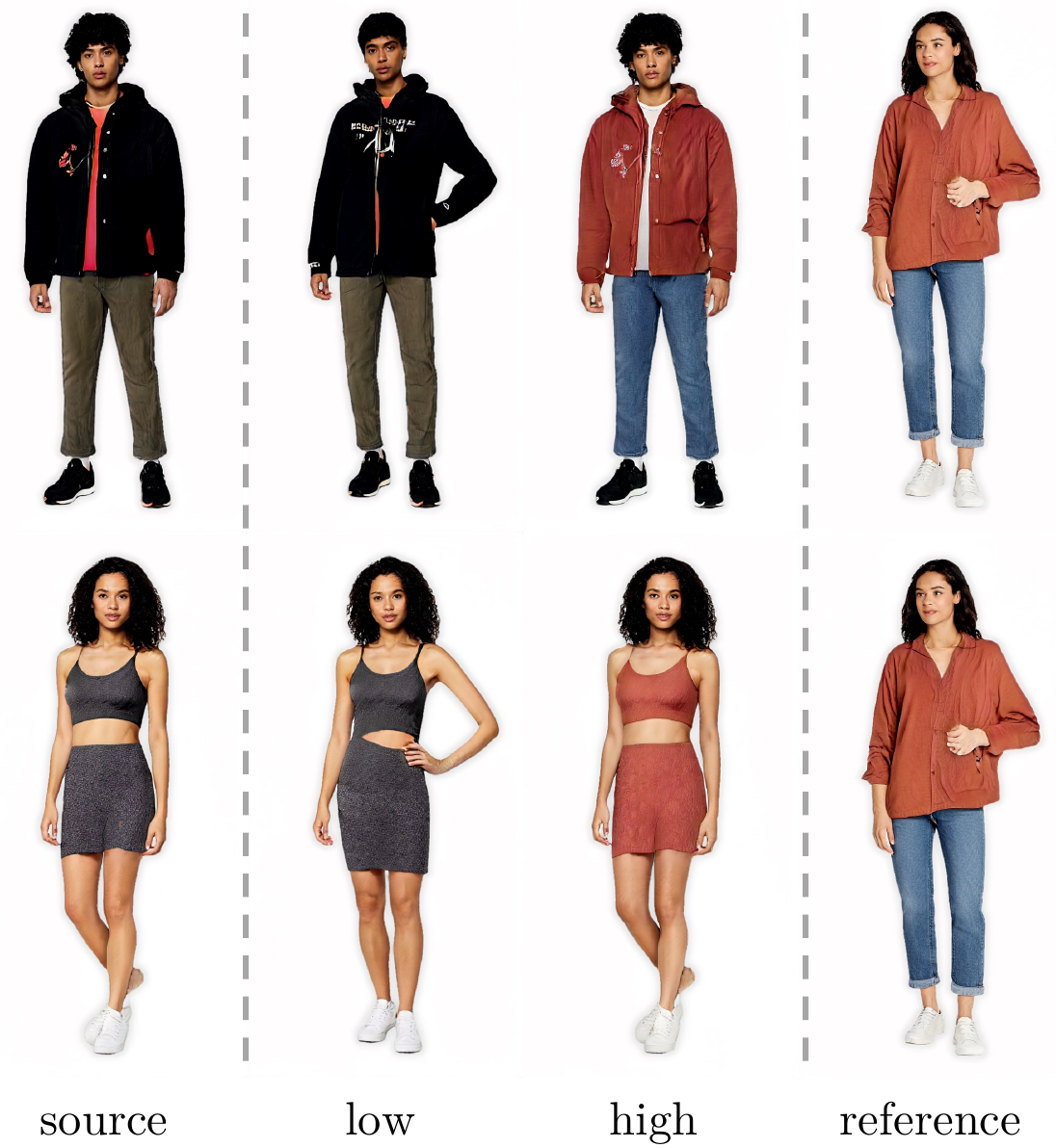}
    % \vspace{-2ex}
    \caption{\textbf{Style mixing with StyleGAN2.} The source and reference images are sampled with StyleGAN2. Style mixing is to replace certain levels of styles of the source with their counterparts from the reference. \textit{Low}-level styles modulate the first few layers of the StyleGAN2 generator while \textit{high}-level styles modulate the last few layers.}
    \label{fig:stylemix}
    % \vspace{-1ex}
\end{figure}

\noindent\textbf{Architecture.} An overview of our generator is presented in Figure \ref{fig:gen}. We use a style-based generator but with a major difference from prior works: different layers of the convolutional backbone are separately modulated by two mapping networks, \ie, a pose mapping network that handles 3D human-body geometry and an appearance mapping network which resembles that of other style-based generators. This design is motivated by the observation that in a StyleGAN2 \cite{stylegan2} model trained on human images. The low-level styles correlate strongly with the pose and orientation of the generated human \cite{fu2022styleganhuman} while others correlate more apparently with the appearance, as shown in Figure \ref{fig:stylemix}. This inspires us to inject the human geometric prior by calculating the low-level styles from the conditioning posed mesh.

\noindent\textbf{3D Pose Mapping Network.} To parse the 3D geometric information, the pose mapping network is formulated as a renderable 3D implicit function $f_p(\psi(\mathbf{x}|\mathbf{p}, \boldsymbol{\theta}), \mathbf{d}) = (\sigma, \mathbf{s}_\mathbf{x})$ where $\mathbf{x} \in \mathbb{R}^3$ is a coordinate in the 3D camera space; $\mathbf{d} \in \mathbb{R}^3$ denotes the orientation of the camera ray; $\mathbf{p} \in \mathbb{R}^{6890 \times 3}$ denotes the coordinates of the body vertices in camera space; $\psi(\mathbf{x}|\mathbf{p}, \boldsymbol{\theta})$ transforms the coordinate $\mathbf{x}$ from the camera space to a canonical space, added to facilitate learning. Following literature in 3D human body reconstruction \cite{zhao2021humannerf,peng2021animatable,mihajlovic2021leap}, $\psi$ is defined as the inverse process of linear blend skinning,
\begin{align*}
\psi(\mathbf{x}|\mathbf{p}, \boldsymbol{\theta}) = {\left ( \sum_{k=1}^{K} \omega_k(\mathbf{x}_\mathbf{p}) \mathbf{G}_k (\boldsymbol{\theta}) \right )} ^ {-1} \mathbf{x}
\end{align*}
$\omega_k(\mathbf{x}, \mathbf{p})$ returns the blend weights corresponding to the vertex nearest to the querying point $\mathbf{x}$; $\mathbf{G}_k (\boldsymbol{\theta}) \in \text{SE}(3)$  denotes the cumulative linear transformation at the $k^{th}$ skeleton joint. The outputs of the implicit function are the opacity $\sigma \in \mathbb{R}$ and a style vector $\mathbf{s} \in \mathbb{R}^{N_s}$, which are used to render a 2D low-resolution pose style map $\mathbf{w}_\mathbf{s} \in \mathbb{R}^{H_s \times W_s \times N_s}$ via ray integration \cite{mildenhall2020nerf}. The implicit function is parameterized as an MLP with periodic activation \cite{sitzmann2020siren} to handle low-dimensional coordinate input.

\noindent\textbf{Neural Rendering.} We use classic volume rendering technique to aggregate predicted style vectors across space. We start by evenly sampling $N$ points $\{ \mathbf{x}_i = \mathbf{o} + t_i \mathbf{d} \}$ within near and far bounds $[t_n, t_f]$ along each camera ray $\mathbf{r}(t) = \mathbf{o} + t \mathbf{d}$. $\mathbf{o}$ denotes the camera center. The style vector on each 2D spatial location is then estimated via

\begin{align*}
\mathbf{S}(\mathbf{r}) & =
\sum_{i=1}^N ~
T_i \left( 1 - \exp \left( - \sigma \left( \mathbf{x}_i \right) \delta_i \right) \right)  \mathbf{s} (\mathbf{x}_i, \mathbf{d}), \\
T_i & = \exp \left(- \sum_{j=1}^{i-1} \sigma(\mathbf{x}_j) \delta_j \right),
\end{align*}
where $\delta_i = \left | \mathbf{x}_{i+1} - \mathbf{x}_{i} \right |$ denotes the distance between adjacent samples.

\noindent\textbf{Appearance Mapping Network}. The appearance mapping network is an MLP $f_a(\mathbf{z})=\mathbf{w} \in \mathbb{R}^{N_w}$, following common practice of style-based generators \cite{stylegan, stylegan2, karras2021alias}. The output style vectors $\mathbf{w}$ do not have spatial dimensions. The final output image is synthesized by the convolutional backbone under modulation from both mapping networks.

\section{Training}

\noindent Our training pipeline is illustrated in Figure \ref{fig:train}. For the network to learn to parse the 3D geometric prior and synthesize images with specified pose, a mapping from 3D geometric information to 2D textures needs to be established. Inspired by \cite{oasis2021}, we use a U-Net \cite{ronneberger2015unet} discriminator architecture together with a segmentation-based GAN loss to establish a one-to-many mapping between geometry and textures. Specifically, the U-Net discriminator classifies each pixel as \textit{fake}, \textit{background} or one of 25 semantic classes (e.g. \textit{head}, \textit{torso} ...), unlike the binary classification on entire images used in traditional GAN training. This approach simultaneously enables free appearance sampling and pose conditioning. A popular practices in pose-conditioned human image generation it to use VAE-style supervision, but the quality of generated image is often outperformed by that of GAN-style training \cite{fruhstuck2022insetgan}. When updating the discriminator, triplets of real images, corresponding SMPL meshes and semantic label maps $(\mathbf{I}, \mathbf{p}_\mathbf{I}, \mathbf{m}_\mathbf{I})$ are needed to calculate the loss.
\begin{align*}
\mathcal{L}^D_\text{adv} = &- \mathbb{E}_{\mathbf{I}} 
\left[ 
\sum_{c}^{N_c} \alpha_c \sum_{x,y}^{H \times W} {\mathbf{m}_\mathbf{I}}_{x,y,c} \log D(\mathbf{I})_{x,y,c}
\right] \\
&- \mathbb{E}_{\mathbf{p},\mathbf{z}} 
\left[ 
\sum_{c}^{N_c} \alpha_c \sum_{x,y}^{H \times W} {\mathbf{m}_\mathbf{0}}_{x,y,c} \log D(G(\mathbf{p}, \mathbf{z}))_{x,y,c}
\right]
\end{align*}
where $x$, $y$ and $c$ are subscripts of the height, width and class dimension. $\alpha_c$ denotes class weights, calculated as the inverse of the per-pixel class frequency. $\mathbf{m}_\mathbf{0}$ denotes a semantic label map in which everywhere is $0$, corresponding to the \textit{fake} class. When updating the generator, pairs of SMPL meshes and semantic label maps $(\mathbf{p}, \mathbf{m}_\mathbf{p})$ are required.
\begin{equation*}
\mathcal{L}^G_\text{adv} = 
- \mathbb{E}_{\mathbf{p},\mathbf{z}}
\left[ 
\sum_{c}^{N_c} \alpha_c \sum_{x,y}^{H \times W}  {\mathbf{m}_\mathbf{p}}_{x,y,c} \log D(G(\mathbf{p}, \mathbf{z}))_{x,y,c}
\right]
\end{equation*}
The segmentation-based GAN loss is sufficient for training the model to generate consistent images with desired pose and view-angle. In practice, we find adding two additional loss terms improves image quality. The first is a perceptual loss \cite{johnson2016perceptual} which minimizes the difference between feature activations extracted from real and generated images using a pretrained VGG \cite{simonyan2014vgg} network.
\begin{equation*}
\mathcal{L}^G_\text{perceptual} = 
\left | \text{VGG}(G(\mathbf{p}_\mathbf{I}, \hat{\mathbf{z}}_\mathbf{I})) - \text{VGG}(\mathbf{I}) \right |
\end{equation*}
where $\hat{\mathbf{z}}_\mathbf{I}$ is an appearance latent code corresponding with the groundtruth image $\mathbf{I}$. These appearance codes are learnable parameters in the model, initialized by inverting a pretrained StyleGAN2 model \cite{fu2022styleganhuman} with e4e \cite{e4e}. 
During training, the collection of appearance latent codes are optimized but anchored by a latent loss.
\begin{equation*}
\mathcal{L}^G_\text{latent} = 
\left | \hat{\mathbf{z}_\mathbf{I}} - \text{e4e}(\mathbf{I}) \right |
\end{equation*}
It is reported that gradient penalty on the discriminator facilitates stable training and convergence \cite{mescheder2018training, stylegan2}. We implement an R1 regularization for our segmentation-based GAN loss:
\begin{equation*}
\mathcal{L}^D_{R1} = \frac{\gamma}{2} \mathbb{E}_\mathbf{I} { \left \| \nabla_\mathbf{I}  D(\mathbf{I}) \right \| }^ 2
\end{equation*}
The total loss for training the generator is the sum of the above terms.
\begin{equation*}
\mathcal{L}^G = \lambda_\text{adv} \mathcal{L}^G_\text{adv} + \lambda_\text{perceptual} \mathcal{L}^G_\text{perceptual} + \lambda_\text{latent} \mathcal{L}^G_\text{latent} + \gamma \mathcal{L}^D_{R1}
\end{equation*}

\section{Preserving Consistency}

\noindent Although capable of producing high-quality results, CNN-based generator networks often sacrifice consistency under geometric transformations \cite{chan2021efficient}. This is a crucial challenge is our approach. Our network is carefully designed to preserve the consistency of appearance against varying pose and view-angle.

\begin{figure}
    \centering
    \includegraphics[width=0.8\linewidth]{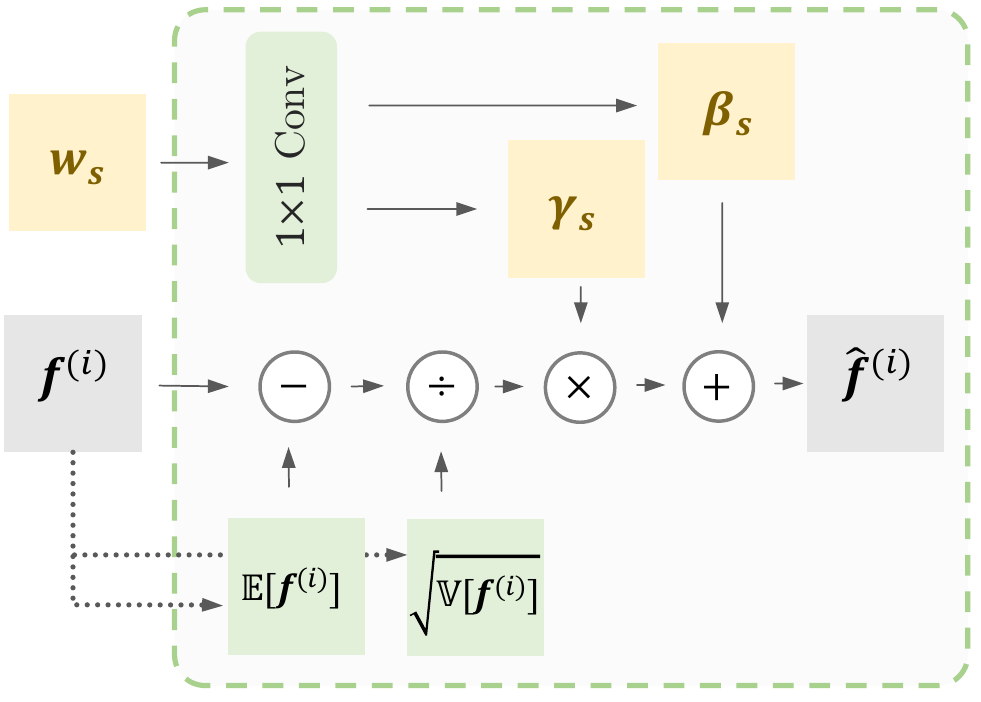}
    \caption{\textbf{Spatial adaptive batch normalization} first performs a statistics-based per-channel normalization on the feature maps and then a re-normalization with mean and variance values calculated from the style maps.}
    \label{fig:san}
    % \vspace{-3ex}
\end{figure}

% we carefully choose our network components and introduce the \textit{spatial adaptive batch normalization} (SABN) which is used to pass the style maps into the backbone in a way that does not impair underlying structure.

\noindent\textbf{Pixel-wise independent convolutional backbone}. Our backbone network is based entirely on $1 \times 1$ convolutions and no upsampling is used. The input to the backbone network is a constant coordinate grid $\mathbf{g} \in \mathbb{R}^{H \times W \times 2}$ containing normalized screen coordinates ranged $[-1, 1]$. Thereby, each pixel is synthesized independent of others. The purpose of this is to eliminate unwanted positional references so that the network is equal-variant to geometric transformations \cite{karras2021alias}. 

\noindent\textbf{Spatial-adaptive batch normalization}. The spatial adaptive batch normalization which is used to pass the style maps into the backbone in a way that does not impair underlying structure. Let us denote by $\mathbf{f}^{(i)}$ the output feature map of the $i^{th}$ layer of the backbone. A spatial adaptive normalization \cite{park2019spade} function $\phi(\mathbf{f}^{(i)}, \mathbf{w}_\mathbf{s})=\hat{\mathbf{f}}^{(i)}$ is needed to inject the style maps rendered by our pose mapping network into the feature maps. As shown in Figure \ref{fig:san}, spatial adaptive normalization first performs a statistics-based per-channel normalization on the feature maps and then a re-normalization with mean and variance values calculated from the style maps. In the first normalization step, how the statistics are calculated makes a crucial difference. When the statistics are calculated on individual feature maps in a mini-batch, \ie instance normalization \cite{ulyanov2016instance}, which is used by \cite{stylegan, park2019spade}, consistency is impaired. We hypothesize that this is because instance normalization removes global affine transformation applied by the previous spatial normalization and forces the normalized layer to resort to controlling only the presence of finer features rather than their precise positions, as similarly observed in \cite{karras2021alias}. 
We propose to use \textit{spatial adaptive batch normalization}:
\begin{equation*}
\phi(\mathbf{f}^{(i)}_{x, y, c}, \mathbf{s}) =
\gamma^{(i)}(\mathbf{s}) _{x, y, c} \cdot
\frac
{\mathbf{f}^{(i)}_{x, y, c} - \mathbb{E}_{\mathbf{p}, \mathbf{z}}[\mathbf{f}^{(i)}_{x, y, c}]}
{\sqrt{\mathbb{V}_{\mathbf{p}, \mathbf{z}}[\mathbf{f}^{(i)}_{x, y, c}]}} 
+ \beta^{(i)}(\mathbf{s})_{x, y, c}
\end{equation*}
where $x$, $y$ and $c$ are the subscripts for the width, height and channel dimension. $\gamma^{(i)}$ and $\beta^{(i)}$ are learned affine transformations. $\mathbb{E}$ and $\mathbb{V}$ estimates the means and variances of the activations of a feature channel across all pixels and all combinations of poses, views and appearances. In practice, this is done by keeping track of the mini-batch statistics of the per-channel activations through exponential moving average. Spatial adaptive batch normalization is used throughout the generation pipeline. For the style vectors $\mathbf{w}$ calculated with the appearance network, the same re-normalization is broadcasted to all spatial locations.

\begin{figure}
    \centering
    \includegraphics[width=\linewidth]{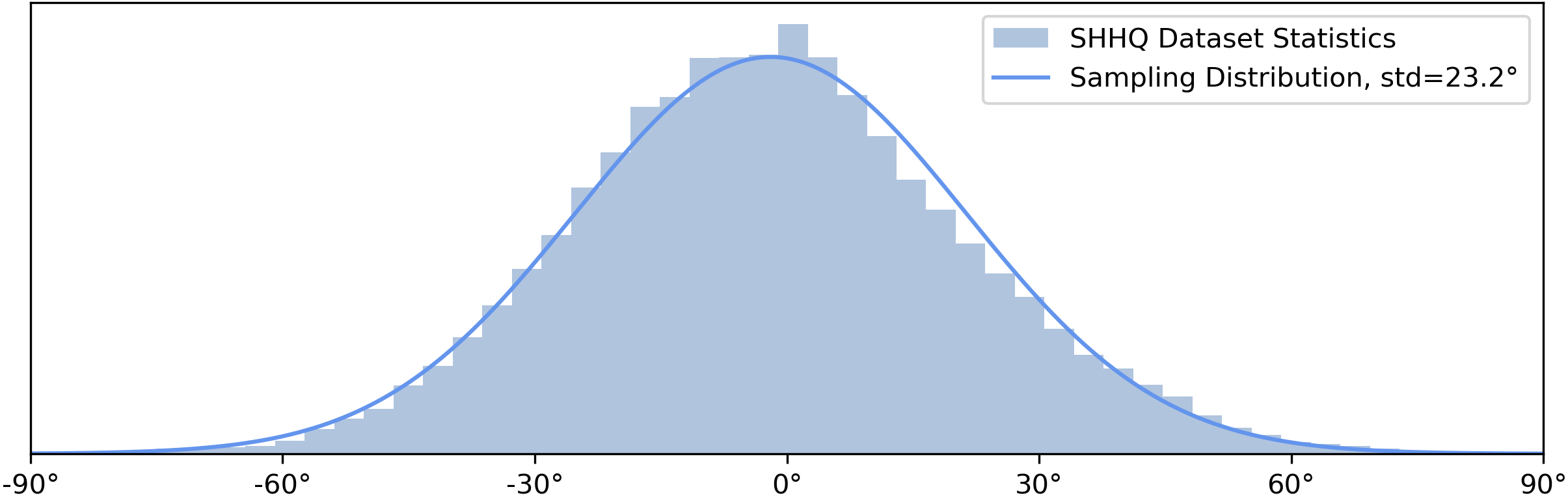}
    % \vspace{-2ex}
    \caption{\textbf{Distribution of view-angle (yaw) in the SHHQ dataset.} We sample view-angle from a normal distribution fitted to the dataset for training and evaluation.}
    \label{fig:shhqhist}
    % \vspace{-4ex}
\end{figure}

\begin{figure*}
    % \vspace{-2ex}
    \centering
    \includegraphics[width=\linewidth]{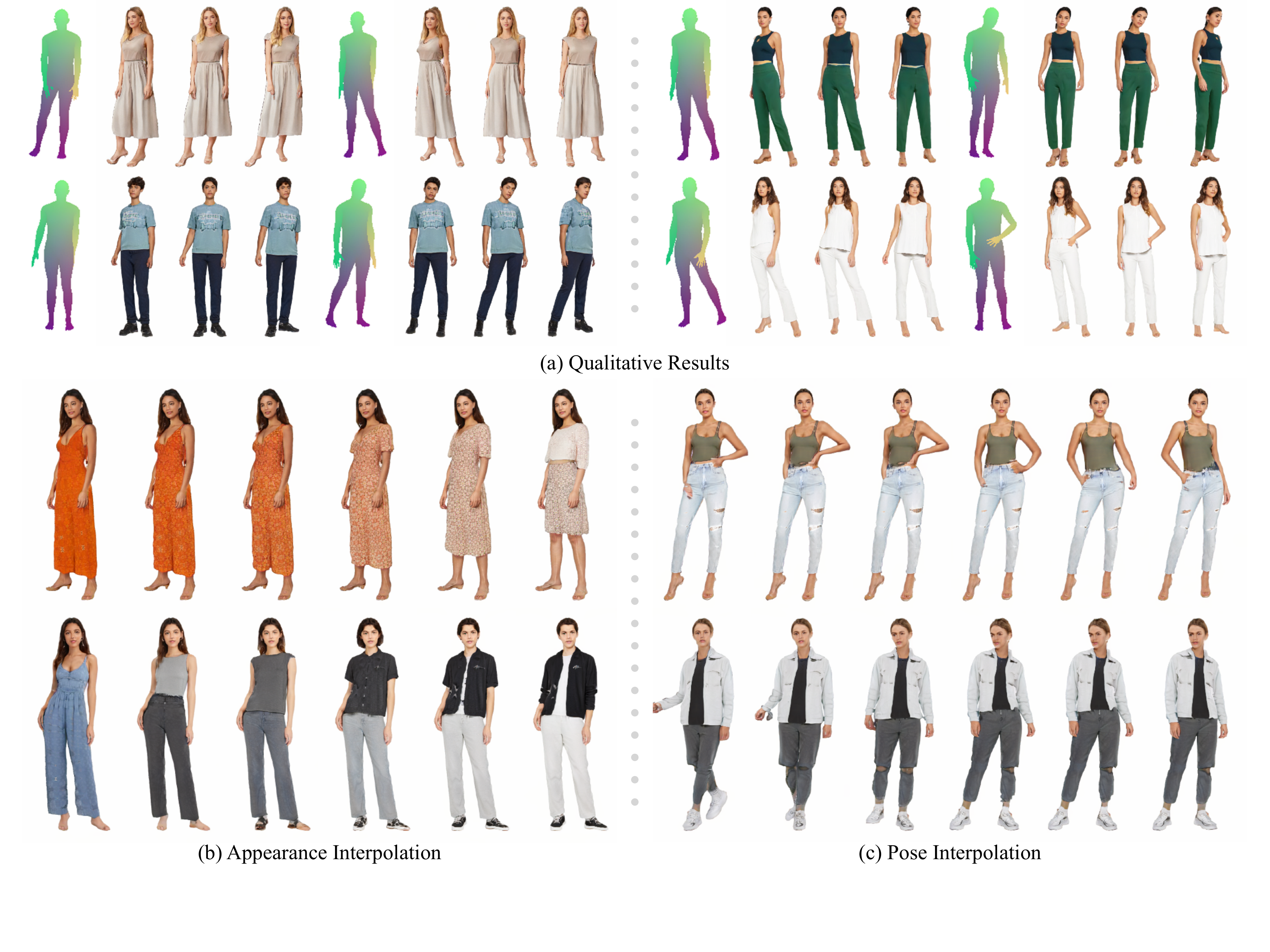}
    \caption{\textbf{Qualitative Results.} In \textbf{(a)}, each row shows two cases separated by the dotted line. For each case we show one identity in two poses and three view-angles. The conditioning mesh is shown on the left of each case. \textbf{(b)} shows two cases of appearance interpolation. \textbf{(c)} shows two cases of pose interpolation.}
    \label{fig:qualitative_ours}
    % \vspace{-1ex}
\end{figure*}

% \begin{figure*}
%     \centering
%     \includegraphics[width=\linewidth]{figures/fig_interp.pdf}
%     \vspace{-2ex}
%     \caption{\textbf{Interpolation.} The left part shows appearance interpolation. The right part shows pose interpolation.
%     }
%     \label{fig:interp}
%     \vspace{-1ex}
% \end{figure*}

\section{Experiments}

\subsection{Dataset} 
We compare 3D-aware human image generation performance on SHHQ \cite{fu2022styleganhuman}, a collection of $\sim220$k real-world full-body human images. All compared models are trained on this dataset. We use an off-the-shelf model PARE \cite{kocabas2021pare} to register a SMPL mesh for each of the training images. The semantic label maps are then obtained by rasterizing the conditioning posed mesh (mapping between mesh vertices and semantic labels is known). We augment the semantic label maps with estimations from a 2D semantic segmentation method DensePose \cite{Guler2018DensePose}. The latter's results are better aligned with the images compared to the estimations from PARE. During training, pose, view and appearance are independently sampled. Poses are directly sampled from data. The view-angles for training and evaluation are sampled from a normal distribution fitted to the dataset i.e. yaw $\sim \mathcal{N}(0, 0.4)$ as shown in Fig. \ref{fig:shhqhist}.

\subsection{Baselines}
We compare with a 2D unconditional method StyleGAN2 \cite{stylegan2}, a 2D conditional method OASIS \cite{oasis2021}, 3D-aware unconditional methods StyleNeRF \cite{gu2021stylenerf} and EG3D \cite{chan2021efficient} and two 3D-aware conditional methods EVA3D \cite{hong2022eva3d} and ENARF-GAN \cite{noguchi2022enarfgan}. Note that we do not compare with fully-3D methods such as \cite{chan2021pi,schwarz2020graf,niemeyer2021giraffe} because these methods do not support training at a meaningful resolution. To evaluate and compare the pose-conditioning capability, we implement pose conditioning for the unconditional methods by first inverting a human image with desired pose to the style-space and then mixing these style vectors with the ones calculated from the randomly sampled $\mathbf{z}$. For StyleGAN2, the inversion method used is e4e\cite{e4e}. For StyleNeRF and EG3D, we train a ResNet \cite{he2016deep} encoder to infer the latent code given camera parameters.

\subsection{Evaluation Metrics}

\noindent \textbf{Fidelity}. We measure fidelity with Fréchet inception distance \cite{heusel2017fid} and kernel inception distance \cite{binkowski2018demystifying}. Two sets of fidelity metrics are calculated for unconditional generation and pose-conditional generation, denoted as (\textit{FID-u}, \textit{KID-u}) and (\textit{FID-p}, \textit{KID-p}), respectively. These values are calculated between $50,000$ real and generated images using an alias-free implementation \cite{parmar2022aliased}. 

\noindent \textbf{Accuracy of Pose Conditioning}. To evaluate the accuracy of generated pose under conditioning, we use PARE \cite{kocabas2021pare} to re-estimate poses from generated images and calculate the 2D mean per-joint position error \cite{ionescu2013human36m} in normalized screen space ranged $[-1, 1]$. The procedure is: i) We randomly sample $4000$ poses from the dataset to serve as ground truth; ii) we use the candidate method to generate $4000$ images conditioned on these poses; iii) the 3D poses of generated persons are estimated using PARE and used to calculate the MPJPE (Protocol $\#1$) \cite{ionescu2013human36m}.

\noindent \textbf{Consistency}. We measure the consistency of appearance under varying pose and view-angle using the signal restoration metric peak signal-to-noise ratio (PSNR) in decibels (dB), following \cite{karras2021alias,zhang2019making}. Specifically , we perform a reverse mapping from pixels to mesh vertices with meshes estimated using \cite{kocabas2021pare}. The PSNR is then calculated between corresponding vertices on these colored meshes. The \textit{Con-v} and \textit{Con-p} respectively denote PSNR calculated under view change and pose change.

\subsection{Qualitative Results}

\noindent Figure \ref{fig:qualitative_ours}a shows 3D-aware human image generation results in multiple poses and view-angles. The appearances of the humans generated with our method is consistent under view and pose changes. Figure \ref{fig:qualitative_ours}b and \ref{fig:qualitative_ours}c show the interpolation results. Our model is able to generate smooth interpolation of appearance and pose.

\begin{figure*}
    \centering
    % \vspace{-1ex}
    \includegraphics[width=\linewidth]{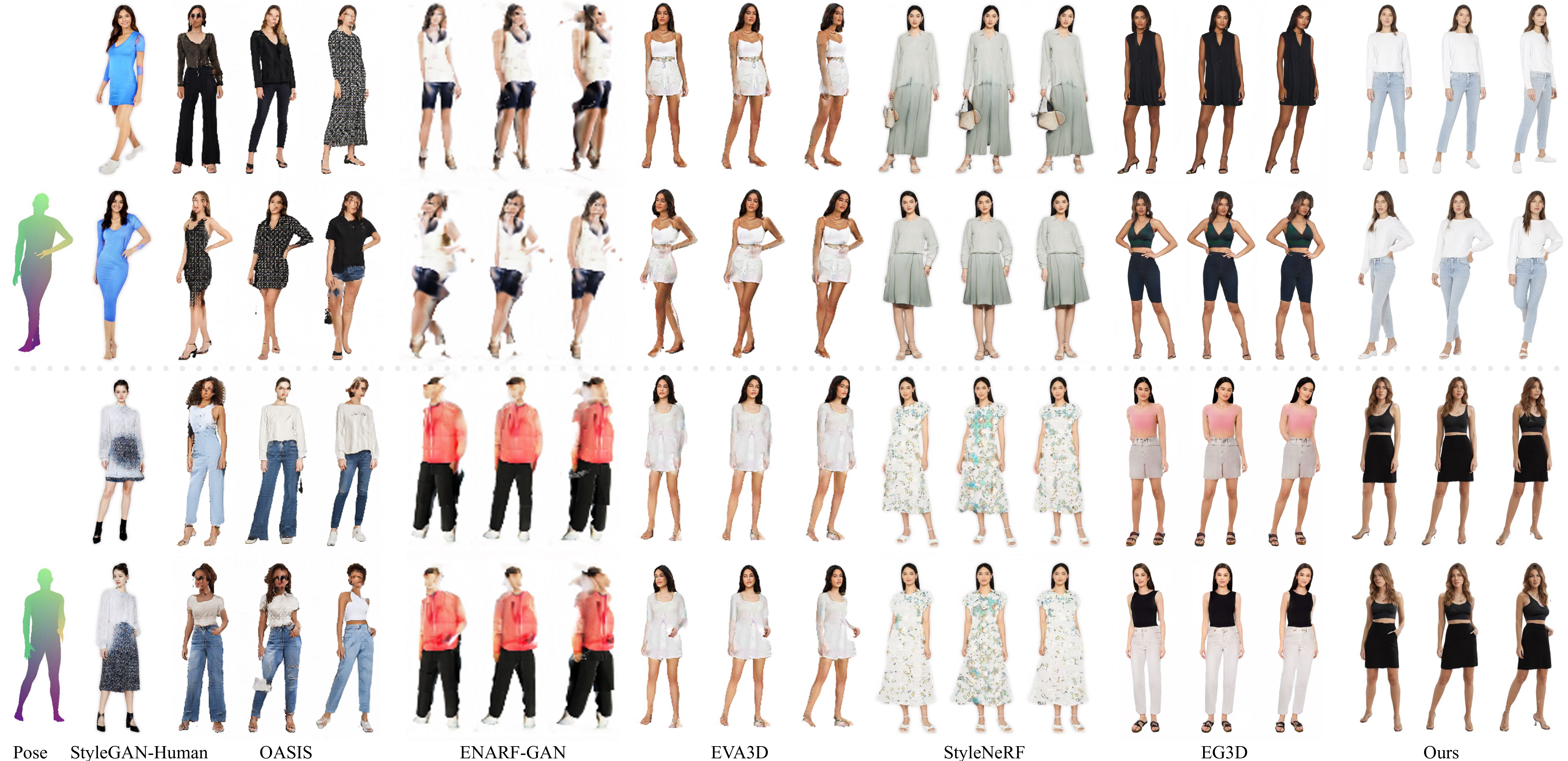}
    \caption{\textbf{Qualitative Comparison.} We show two cases separated by the dotted line. For each case, the first row shows unconditional generation results, the second row shows pose-conditional generation results. We show three view angles, from $-30^{\circ}$ to $30^{\circ}$.
    }
    \label{fig:qualitative}
    % \vspace{-1ex}
\end{figure*}

\begin{table}
\caption{\textbf{Quantitative Comparisons} on human images generated by 2D and 3D methods. \textit{KID} values are reported in $10^{-3}$ units; \textit{Pose} values are reported in $10^{-2}$ units. The first place and runner-up in each metric are reported in \textbf{bold} typeface. Our method achieves the lowest fidelity scores and the highest score of appearance consistency under different body-poses.}
% \vspace{-1ex}
\begin{center}

\resizebox{\linewidth}{!}{
\begin{tabular}{l|cccccc|cc}
\multirow{2}{*}{Metrics} & 
\multirow{2}{*}{\begin{tabular}[c]{@{}l@{}}StyleGAN-\\ Human, 512\end{tabular}} & 
\multirow{2}{*}{\begin{tabular}[c]{@{}l@{}}OASIS,\\ 256\end{tabular}} & 
\multirow{2}{*}{\begin{tabular}[c]{@{}l@{}}ENARF-\\ GAN, 128\end{tabular}} & 
\multirow{2}{*}{\begin{tabular}[c]{@{}l@{}}EVA3D,\\ 256\end{tabular}} & 
\multirow{2}{*}{\begin{tabular}[c]{@{}l@{}}StyleNeRF,\\512\end{tabular}} & 
\multirow{2}{*}{\begin{tabular}[c]{@{}l@{}}EG3D,\\ 512\end{tabular}} & 
\multirow{2}{*}{\begin{tabular}[c]{@{}l@{}}Ours,\\ 256\end{tabular}} & 
\multirow{2}{*}{\begin{tabular}[c]{@{}l@{}}Ours,\\ 512\end{tabular}} \\
& \multicolumn{1}{l}{} & \multicolumn{1}{l}{} & \multicolumn{1}{l}{}& \multicolumn{1}{l}{} & \multicolumn{1}{l}{}& \multicolumn{1}{l}{} & \multicolumn{1}{|l}{}& \multicolumn{1}{l}{} \\
\Xhline{1.2pt}
FID-p $\downarrow$  & \textbf{6.68} & 10.74 & 30.40 & 26.91 & 13.13 & 17.54 & \textbf{6.61}   & 9.31 \\
KID-p $\downarrow$  & \textbf{3.60} & 8.82  & 23.66 & 18.71 & 7.01  & 10.71 & \textbf{3.01}   & 5.16 \\
FID-u $\downarrow$  & \textbf{2.83} & - & - & - & \textbf{7.60}   & 13.05 & -& -    \\
KID-u $\downarrow$ & \textbf{1.50} & - & - & - & \textbf{3.96}  & 8.36 & -& -    \\
\hline
Pose $\downarrow$  & 4.48 & \textbf{1.82} & 6.66 & 3.12 & 6.64  & 8.39 & 2.21 & \textbf{2.08} \\
\hline
% Con-v & - & - & 20.88 & 21.98 & \textbf{24.45} & 23.00 & \textbf{24.00} \\
% Con-p & 17.13 & 17.04 & 15.82 & 19.07 & 21.34 & 18.03 & 19.01 \\
Con-v $\uparrow$ & -    & 15.68 & 19.39 & \textbf{25.95} & 18.29 & 19.93 & 21.83 & \textbf{22.84}  \\
Con-p $\uparrow$ & 12.77 & 13.19 & 15.67 & \textbf{19.16} & 13.74 & 11.91 & 17.72 & \textbf{18.93} \\
\Xhline{1.2pt}
\end{tabular}
}

\end{center}
% \vspace{-2ex}
\label{table:sota}
\end{table}

\begin{table}
\caption{\textbf{Ablation study} quantitative results. \textit{KID} values are reported in $10^{-3}$ units; \textit{Pose} values are reported in $10^{-2}$ units. The first place and runner-up in each metric are reported in \textbf{bold} typeface.}
\begin{center}
% \vspace{-1ex}
\resizebox{\linewidth}{!}{
\small
\begin{tabular}{l|ccccc}
Config. & FID-p $\downarrow$ & KID-p $\downarrow$ & Pose $\downarrow$    & Con-v $\uparrow$  & Con-p $\uparrow$  \\
 % & & \times 10^{-3}\times 10^{-3} & \times 10^{-2}\times 10^{-2} & & \\
\Xhline{1.2pt}
VAE  & 8.00          & 4.02          & 9.54          & 19.22          & 15.18          \\
FF   & 9.42          & 4.57          & 2.60          & 21.63          & 17.89 \\
Up   & 8.28          & 5.42          & \textbf{1.89} & 16.03          & 13.16          \\
IN   & 8.90          & 4.04          & 1.93          & 17.73          & 13.36          \\
\hline
$\text{Full}_\text{mixed}$ & \textbf{7.98} & \textbf{3.95} & 2.46 & \textbf{24.80} & \textbf{18.54} \\
$\text{Full}_\text{all}$ & 8.15 & 4.25 & 2.28 & \textbf{22.23} & \textbf{18.60} \\
$\text{Full}$ & \textbf{6.61} & \textbf{3.01} & \textbf{2.21}  & 21.83 & 17.72 \\
\Xhline{1.2pt}
\end{tabular}
% \begin{tabular}{l|cccc|c}
% Metrics & VAE   & Feed Forward             & Upsampling            & Instance Norm    & Full           \\
% \Xhline{1.2pt}
% FID-p   & \textbf{8.00}  & 9.42           & 8.28          & 8.90  & \textbf{6.61}  \\
% KID-p   & \textbf{4.02}  & 4.57           & 5.42          & 4.04  & \textbf{3.01}  \\
% \hline
% Pose    & 9.54  & 2.60           & \textbf{1.89} & \textbf{1.93}  & 2.21           \\
% \hline
% Con-v   & 19.22 & \textbf{21.63}          & 16.03        & 17.62 & \textbf{21.83} \\
% % Con-p   & 10.13 & 8.81           & 8.19         & 8.33  & 8.56           \\
% % Con-v   & 19.22 & 21.63          & 16.03         & 17.73 & \textbf{21.83} \\
% Con-p   & 15.18 & \textbf{17.89} & 13.16         & 13.57 & \textbf{17.72} \\
% \Xhline{1.2pt}
% \end{tabular}
}
\end{center}
% \vspace{-1ex}
\label{table:ablation}
\end{table}

\subsection{Comparisons}

\noindent Quantitative results are presented in Table \ref{table:sota}. Qualitative results are shown in Figure \ref{fig:qualitative}. Our method demonstrates the best overall performance, with an FID comparable to the 2D state-of-the-art method StyleGAN-Human, competitive pose conditioning accuracy and top-level consistency. Using GAN inversion methods, we can produce images with StyleGAN2, StyleNeRF and EG3D that match the condition pose closely but not exactly. This suggests that the representations learned by these methods are entangled in appearance and pose. Style mixing is not enough to achieve clean separation in the two domains. Our method is able to generate images that accurately matches the conditioning pose. Relative to leading 3D-aware, pose-conditional approaches like EVA3D and ENARF-GAN, our technique exhibits superior capability in producing high-fidility images. EVA3D showcases remarkable consistency, it is plausible that their single-stage rendering framework, circumventing any upsampling or the use of any 2D convolutional network, contributes to this performance. However, this choice comes at a cost of high GPU memory consumption. Besides, all of the compared baselines except EVA3D show different extents of inconsistencies under view and pose variation. This is another indicator of entanglement of appearance and pose in the learned latent spaces of the baselines. In contrast, the appearance of the humans generated by our method remains consistent under pose and view variation.  
% the unconditional StyleGAN2 and StyleNeRF are able to generate images with target pose close to the condition.
% However, the high-dimensional latent space is complicated, where using a single style mapping network can not disentangle the appearance and pose styles well.
% s somewhat close to the conditioning image.
% OASIS shows the best pose-conditioning accuracy and outperforms our method in this metric, but performs poorly in image consistency.

\subsection{Ablation Study}
% TODO: refine
% To assess the effectiveness of our contributions, we evaluate three models, each of which ablates a key part of our design while other settings remain untouched. 
We compare our 3d-aware generator with different ablated models that use previous methods instead of our designed modules to measure the effectiveness of each module. Specifically,
to examine the effects of passing the 3D geometric prior into the 2D backbone as styles, i.e. the 2D-3D hybrid generator, we compare with the feed-forward approach used by \cite{gu2021stylenerf, chan2021efficient, zhou2021CIPS3D}, denoted as (\textit{Feed Forward}). 
To evaluate the contribution of the segmentation-based GAN loss, we train a model in which this loss is replaced with the traditional GAN loss with binary discrimination. This results in a model trained under VAE-style supervision as in \cite{hong2021headnerf}, denoted as \textit{VAE}.
To examine the effectiveness of the pixel-independent backbone design, we compare with a variant that uses traditional upsampling convolution network like in \cite{oasis2021, stylegan2}, which is denoted as \textit{Upsampling}. 
To examine the effectiveness of spatial adaptive batch normalization, we compare with a variant that uses spatial adaptive instance normalization instead, denoted as \textit{Instance Norm}. 

Based on different ways of injecting the styles into the convolutional backbone, we derive three variants of the full model. The default model, \textit{Full}, introduces pose style maps $\mathbf{w}\textbf{p}$ into the initial three layers and appearance styles $\mathbf{w}$ into the subsequent six. $\textit{Full}_\text{all}$ adds the elementwise sum of $\mathbf{w}\textbf{p}$ and $\mathbf{w}$ to every layer, while $\textit{Full}_\text{mixed}$ incorporates $\mathbf{w}_\textbf{p}$ into the first three and $\mathbf{w}$ throughout all layers.

All models in this experiment are trained and evaluated in the resolution of $256 \times 128$. Quantitative results are presented in Table \ref{table:ablation} and qualitative results are presented in the supplementary materials. Our full models perform best in fidelity and view-consistency. The \textit{Feed-Forward} demonstrates favorable consistency, but with no small sacrifice in image quality. In other terms, passing the geometric prior in the form of low-level styles has a small toll on pose-accuracy and consistency, but completely within the acceptable range. Meanwhile, the image quality is greatly improved. By examining the results of the \textit{VAE} setting, we can see that segmentation-based GAN loss is crucial to pose accuracy. It also improves fidelity, possibly because this loss helps the model to make better use of the 3D Human Prior. The \textit{Upsampling} and \textit{Instance Norm} configurations show the best accuracy of generated pose but with impaired image consistency in different extents. These two configurations show that the pixel-wise independent backbone and spatial adaptive batch normalization indeed preserves consistency. Finally, the variants based on different ways of style injection shows that is beneficial to inject the pose styles into only the initial layers of the network.

\begin{figure}
    \centering
    \includegraphics[width=\linewidth]{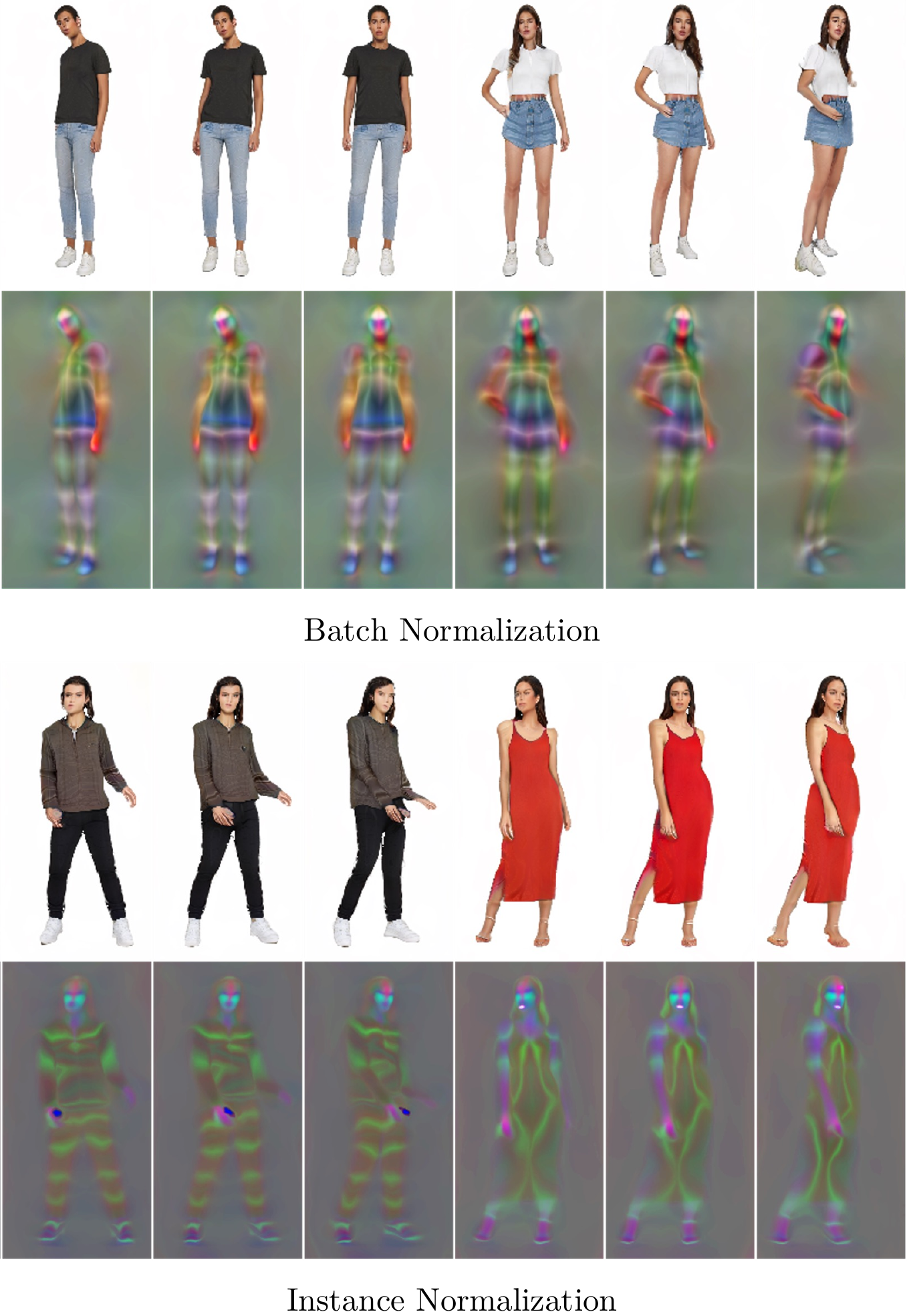}
    % \vspace{-1ex}
    \caption{\textbf{Internal Representation Visualization.} Two identities in two different poses are visualized for each model. For the same model, the same feature channel are shown for the two identities.}
    \label{fig:internal}
    % \vspace{-1ex}
\end{figure}

\subsection{Internal Representation}

\noindent Figure \ref{fig:internal} visualizes typical internal representations from our networks. One interesting obervation is that in the model that uses batch normaliztion, we are able to find certain feature activations that are consistent across different poses, view-angles and appearances, such as the ones depicted. However, such activations are absent in the model with instance normalization. This may be because the instance normalization model mainly modulates the existence of image features rather than their location, as also noted by \cite{karras2021alias}. We hypothesize that batch normalization preserves an internal coordinate system that helps map different image textures to the 3D surface, which improves image consistency.

\section{Implementation Details}
\label{section:implementation_details}

\noindent \textbf{Training Details.} Since both photometric (perceptual loss) and adversarial supervision (segmentation-based GAN loss) are used when training our model, we design two different types of training iterations, \ie partially-conditional iterations and conditional iterations, which are executed alternately. For partially-conditional iterations, we sample appearance codes, poses and view-angles independently. The appearance codes are sampled from a normal distribution $\mathbf{z} \sim \mathcal{N}(0, 1)$. The poses are randomly sampled from all training data. The view-angles are sampled from a uniform distribution $\mathbf{p} \sim \mathcal{N}(0, 0.4)$ based on training data statistics. Only the segmentation-based GAN loss and the $R1$ regularization are calculated for this kind of iteration. For conditional iterations, we sample quadruplets of image, appearance code, pose and view-angle $(\mathbf{I}, \hat{\mathbf{z}_\mathbf{I}}, \mathbf{p}_\mathbf{I}, \mathbf{v}_\mathbf{I})$. The appearance codes $\hat{\mathbf{z}_\mathbf{I}}$ are sampled from the code pool $\mathbf{Z} \in \mathbb{R}^{N_\mathbf{I} \times N_z}$ which is a trainable parameter of the model. It contains $N_\mathbf{I}$ entries of appearance codes, each corresponds with an image in the training dataset, initialized by inverting a pretrained StyleGAN2 model \cite{fu2022styleganhuman} with e4e \cite{e4e}. The pose $\mathbf{p}_\mathbf{I}$ and view-angle $\mathbf{v}_\mathbf{I}$ are estimated using PARE \cite{kocabas2021pare} Both the perceptual loss and the segmentation-based GAN loss are calculated for conditional iterations. When calculating the segmentation-based GAN loss, we augment the ground truth semantic maps that are calculated with PARE \cite{kocabas2021pare} estimation results with the ones estimated by DensePose \cite{Guler2018DensePose} with a $50\%$ probability. This is because the results from PARE does not align with the input image as well as the results from DensePose.

\noindent \textbf{Hyperparameters.} Our models are trained $300$k steps using an Adam \cite{kingma2014adam} optimizer. The learning rate is set as $0.0001$ for the generator backbone and $0.0004$ for the discriminator. The learning rate of the two mapping networks is set as $5 \times 10^{-6}$. The batch size is set as $32$. Spectral normalization \cite{miyato2018spectral} on model parameters is used to improve training stability. The weights of the generator's loss terms are $\lambda_\text{adv}=1$, $\lambda_\text{percept}=1$, $\lambda_\text{latent}=0.05$ and $\gamma=0.25$. After training for $130k$ steps, learning rates are decreased by half; $\lambda_\text{percept}$ and $\lambda_\text{latent}$ are set as $0$. We use lazy execution of $R1$ regularization to speedup training. The $R1$ term is updated once every $4$ iterations. Our $256 \times 128$ resolution model takes $3$ days to train on $4$ A100 GPUs. Our $512 \times 256$ model requires $5$ days on $8$ A100 GPUs.

\noindent \textbf{Inversion-based baselines} 
For unconditional generative models like StyleGAN-Human, StyleNeRF and EG3D, we follow the inversion-and-manipulation manner to explicitly control the pose/view of the human image. Specifically, for StyleGAN-Human, we adopt the pre-trained e4e~\cite{e4e} model to invert the source and target images to $w_s$ and $w_t$, respectively. We then replace the injection style $w_s$ in the first six synthesis layers with $w_t$ to control the pose and global orientation of the human. 
3D-aware generative models require the extrinsic camera as a conditional input to control the image's view. Therefore, we construct an encoder consisting of $\log_{2}{N}$ ResBlock layers and a fully connected layer to predict the style latent $w$ and camera parameters $c$.  We adopt a self-supervised learning scheme by using synthesized images as input and corresponding style latent $w$ and random camera $c$ as ground-truth. After training, we can edit the pose/view of the source image by replacing the injection style in the first six synthesis layers with the predicted latent (as in StyleGAN-Human) and using predicted camera as condition.

\section{Limitations}

Our model can handle most of the view-angles and poses in the training data, but it fails to generalize beyond them. Exploring better generalization and extrapolation methods could be a promising direction for future work. We also notice that inconsistencies in the appearance are still observable in some cases when the view-angle changes. We believe this can be improved by closely inspecting the CNN backbone from the perspective of signal processing to ensure equivariance. Finally, since our framework uses a disentangled 3D representation for pose, efficient scene-specific 3D representations \cite{muller2022instant} could be employed for better computational performance.

\section{Conclusion}

We build a generative adversarial networks 3DHumanGAN that synthesizes full-body human images with photorealistic 3D-awareness. Our generator combines 3D geometric prior of the human body with a 2D-3D hybrid structure and produces consistent appearance across different poses and view-angles with high image quality. Our segmentation-based GAN loss is essential for guiding the generator to parse and condition on the 3D human body prior. We demonstrate through an ablation study and internal representation visualization that our pixel-independent backbone and spatial adaptive batch normalization technique effectively preserve consistency.

{\small
\bibliographystyle{ieee_fullname}
\bibliography{references}
}

% \clearpage
% \appendix

\end{document}

%%%%%%%%% TITLE
% \title{\ Unsupervised Learning to Dance by Surfing The Internet}

% \title{\ Self-Taught Dancer: Unsupervised Motion Retargeting via \\Representation Disentanglement}
\title{3DHumanGAN: 3D-Aware Human Image Generation with 3D Pose Mapping\\[5pt]
{\large Supplementary Material}}

\author{Zhuoqian Yang$^{1, 2 \dagger}$ \qquad Shikai Li$^{1}$ \qquad Wayne Wu\textsuperscript{1 \Letter} \qquad Bo Dai$^{1}$  \\
$^{1}$ Shanghai AI Laboratory \hspace{10pt}
$^{2}$ School of Computer and Communication Sciences, EPFL \\
% \vspace{-5mm}
{\tt\small zhuoqian.yang@epfl.ch}\qquad
{\tt\small lishikai@pjlab.org.cn}\qquad
\\
{\tt\small wuwenyan0503@gmail.com}\qquad
{\tt\small daibo@pjlab.org.cn}
% \vspace{-10mm}
}

\maketitle

\blfootnote{
$\dagger$ Work donw as research engineer at Shanghai AI Laboratory. 
}

\begin{abstract}

This document provides supplementary information which is not elaborated in our manuscript due to space limits. Section \ref{section:consistency} discusses the relation between consistency and equivariance and the reasons for using spatial adpative batch normalization from a theortical angle. 
% Section \ref{section:implementation_details} presents details about the implementation of our method. 
Section \ref{section:qualitative_ablation} presents qualitative results for the ablation study. Section \ref{section:additional_qualitative} presents additional qualitative results for our method and comparison against prior art. 
% Section \ref{sec:3} presents additional qualitative results.
We also present a video demo which includes a brief introduction of our method and animated qualitative results.
\end{abstract}

\section{Consistency and Equivariance}
\label{section:consistency}

We have mentioned in the main text that equivariance is essential for producing consistent outcomes. Here we elaborate on this point. We employ a convolutional backbone that consists only of $1\times1$ convolutions. The backbone takes a 2D grid of image coordinates as input. In this setting, we hypothesize that the 3D pose mapping network performs certain non-linear spatial transformations on the 2D coordinates based on the SMPL mesh that conditions it. To ensure that these transformations are reflected in the generated image, we need a CNN backbone that is equivariant. Mathematically, a function is equivariant if applying a transformation to its input leads to the same result as applying the transformation to its output. We think this is vital for the network to learn features that are not tied to absolute coordinates but follow the coordinates transformed under 3D guidance.

In the main text we compared between spatial adaptive instance normalization (SAIN) and spatial adaptive batch normalization (SABN) for injecting the style maps rendered by the 3D pose mapping network into the 2D backbone. Both of these operations contain a normalization step and a denormalization step. Here we inspect the normalization step in detail. The former uses instance statistics for this step while the latter uses batch statistics. We will use the following notation: $\mathbf{x}$ is an input vector of size $n$, $\mathbf{y}$ is an output vector of size $n$, $\mathbf{W}$ is a weight matrix of size $m \times n$, and $\mathbf{b}$ is a bias vector of size $m$. 

First, let us consider instance normalization. For each instance $\mathbf{x}_i$ in the batch, we have
$$
\mathbf{y}_i = \frac{\mathbf{W}\mathbf{x}_i + \mathbf{b} - \boldsymbol{mu}_i}{\boldsymbol{\sigma}_i},
$$
where $\boldsymbol{mu}_i$ and $\boldsymbol{\sigma}_i$ are the mean and standard deviation of $\mathbf{W}\mathbf{x}_i + \mathbf{b}$. This means that IN maps all instances to have zero mean and unit variance. However, this also means that IN discards some information about the relative magnitude and scale of each instance. When $\mathbf{x}$ store coordinates, instance normalization effectively scales and shifts $\mathbf{x}$ which makes the network unequivariant.

Next, let us consider normalization with batch statistics. For a batch of size $B$, we have
$$
\mathbf{y} = \frac{\mathbf{W}\mathbf{x} + \mathbf{b} - \boldsymbol{\mu}}{\boldsymbol{\sigma}},
$$
where $\boldsymbol{\mu}$ and $\boldsymbol{\sigma}$ are the mean and standard deviation vectors across the batch. When the batch size approaches infinity, we have
\begin{align*}
\lim_{B\to\infty} \boldsymbol{\mu} &= \mathbb{E}[\mathbf{W}\mathbf{x} + \mathbf{b}],\\
\lim_{B\to\infty} \boldsymbol{\sigma}^2 &= \text{diag}(\mathbb{V}[\mathbf{W}\mathbf{x} + \mathbf{b}]),
\end{align*}
where $\mathbb{E}[\cdot]$ denotes expectation and $\mathbb{V}[\cdot]$ denotes covariance. These limits are independent of any particular instance in the batch. Assuming that the entries in $\mathbf{x}$ are independent and normally distributed, \ie $\mathbf{x}_i\sim\mathcal{N}(\mu_x, \sigma^2_x)$, 
\begin{align*}
\lim_{B\to\infty} \boldsymbol{\mu} 
&= \mathbb{E}[\mathbf{W}] \cdot \mathbb{E}[\mathbf{x}] + \mathbb{E}[\mathbf{b}]\\
&= \mu_x \mathbf{W} \cdot  \boldsymbol{1} + \mathbf{b}, \\
\lim_{B\to\infty} \boldsymbol{\sigma}^2 &= \text{diag}(\mathbb{V}[\mathbf{W}\mathbf{x}]) \\
&= \text{diag}(\mathbf{W}^T\mathbb{V}[\mathbf{x}]\mathbf{W})
\\ 
&= \sigma^2_x \text{diag}(\mathbf{W}^T\mathbf{W})
\end{align*}
This normalization maps all instances to have the same mean and variance vectors as the whole batch. However, this does not mean that it discards any information about the relative magnitude and scale of each instance, since these limits only depend on the global statistics of $\mathbf{x}$.
When $\mu_x=0$ and $\sigma_x=1$ the normalization becomes equivariant.

The reasoning above provide mathematical intuitions in preferring SABN over SAIN for better equivariance. However, the equivatiance of SABN still depends on several assumptions. Devloping an operation that do not rely on these assumptions could be a promising direction for future work.

\section{Ablation Study (Qualitative Results)}
\label{section:qualitative_ablation}

\noindent We present qualitative results of the ablation study in Fig. \ref{fig:qualitative_ablation}. Note that all ablation models are trained at $256 \times 128$ resolution due to limited computational resources. Replacing the segmentation-based GAN loss with traditional binary GAN loss causes the model to lose the ability of pose conditioning, as shown by the result of the \textit{VAE} configuration. The model that combines 2D and 3D networks in a feed-forward manner is functional, but less desirable in image quality and consistency. Using an upsampling convolutional backbone instead of pixel-wise independent one results in impaired consistency. When passing the 3D style maps into the 2D backbone, using instance normalization instead of batch normalization has similar effects.

\begin{figure*}
    \vspace{2ex}
    \centering
    \includegraphics[width=\linewidth]{figures/fig_qualitative_ablation.pdf}
    \caption{\textbf{Ablation Study Qualitative Results.} We show four cases separated by dotted lines. For each case we show one identity in two poses and three view-angles. The conditioning mesh is shown on the left of each case.}
    \label{fig:qualitative_ablation}
    \vspace{2ex}
\end{figure*}

\section{Additional Qualitative Results}
\label{section:additional_qualitative}

\begin{figure*}
    \vspace{-2ex}
    \centering
    \includegraphics[width=\linewidth]{figures/fig_qualitative_additional.pdf}
    \caption{\textbf{Additional Qualitative Comparison.} We show four cases separated by the dotted line. For each case, the first row shows unconditional generation results, the second row shows pose-conditional generation results. We show three view angles, from $-30^{\circ}$ to $30^{\circ}$}
    \label{fig:qualitative_additional}
    \vspace{-1ex}
\end{figure*}

We show additional qualitative comparison results in Fig. \ref{fig:qualitative_additional}. For our method, we provide additional qualitative results in Fig. \ref{fig:view_and_pose}, appearance interpolation results in Fig. \ref{fig:appearance_interp} and pose interpolation results in Fig. \ref{fig:pose_interp}. We also provide animated results in the video demo.

\begin{figure*}
    \vspace{-2ex}
    \centering
    \includegraphics[width=0.95\linewidth]{figures/fig_qualitative_additional_view_and_pose.pdf}
    \caption{\textbf{Additional Qualitative Results.} Each row shows two cases separated by the dotted line. For each case we show one identity in two poses and three view-angles. The conditioning mesh is shown on the left of each case.}
    \label{fig:view_and_pose}
    \vspace{-1ex}
\end{figure*}

\begin{figure*}
    \vspace{-2ex}
    \centering
    \includegraphics[width=0.95\linewidth]{figures/fig_qualitative_additional_appearance_interp.pdf}
    \caption{\textbf{Additional Appearance Interpolation Results.} Each row shows two cases separated by the dotted line.}
    \label{fig:appearance_interp}
    \vspace{-1ex}
\end{figure*}

\begin{figure*}
    \vspace{-2ex}
    \centering
    \includegraphics[width=\linewidth]{figures/fig_qualitative_additional_pose_interp.pdf}
    \caption{\textbf{Additional Pose Interpolation Results.} Each row shows two cases separated by the dotted line.}
    \label{fig:pose_interp}
    \vspace{-1ex}
\end{figure*}
% {\small
% \bibliographystyle{ieee_fullname}
% \bibliography{references}
% }